\documentclass[11pt,a4paper]{article}
\usepackage[hyperref]{acl2019}
\usepackage{times}
\usepackage{latexsym}

\usepackage{url}
\usepackage{booktabs}
\usepackage{graphicx}
\usepackage{multirow}
\usepackage{siunitx}
\usepackage{etoolbox}
\robustify\bfseries

\aclfinalcopy 

\title{Measuring the compositionality of noun-noun compounds over time}

\author{Prajit Dhar \\
  Leiden University \\
  \texttt{dharp@liacs.leidenuniv.nl} \\\And
  Janis Pagel \\
  University of Stuttgart \\
  \texttt{pageljs@ims.uni-stuttgart.de} \\\AND
  Lonneke van der Plas \\
  University of Malta \\
  \texttt{lonneke.vanderplas@um.edu.mt}}

\date{}

\begin{document}
\maketitle
\begin{abstract}
We present work in progress on the temporal progression of compositionality in noun-noun compounds.
Previous work has proposed computational methods for determining the compositionality of compounds. These methods try to automatically determine how transparent the meaning of the compound as a whole is with respect to the meaning of its parts. We hypothesize that such a property might change over time. We use the time-stamped Google Books corpus for our diachronic investigations, and first examine whether the vector-based semantic spaces extracted from this corpus are able to predict compositionality ratings, despite their inherent limitations. We find that using temporal information helps predicting the ratings, although correlation with the ratings is lower than reported for other corpora. Finally, we show changes in compositionality over time for a selection of compounds.
\end{abstract}

\section{Introduction}

Compositionality is a long debated issue in theoretical linguistics. The principle of compositionality \cite{partee:84} states that the meaning of an expression is a function of the meanings of its parts and of the way they are syntactically combined.
It is often used to describe how the meaning of a sentence can be derived from the meaning of single words and phrases, but the principle might also be postulated for \textit{compounding}, i.e. the process of combining two or more lexemes to form a new concept \cite[p. 1 and 4]{bauer2017}. Compounds can often be directly derived from the meanings of the involved compound constituents (e.g. \textit{graduate student}, \textit{speed limit}), however, we also find compounds whose meanings can only be derived partially from their components (\textit{night owl}, \textit{hot dog}).

Surprisingly, diachronic perspectives on compositionality\footnote{A notable exception is \citet{vincent2014}, although he mainly focuses on syntactic processes in Romance languages and only briefly covers numeral words.} are virtually absent from previous work. To the best of our knowledge, we present the first study on the compositionality of compounds over time. We bring two strands of research together. On the one hand we are inspired by the synchronic work on predicting the degree of compositionality of compounds by comparing the vector-based representations of the parts to the vector-based representations of the compound as a whole. On the other hand, we rely on methods designed for detecting semantic change, such as presented in \citet{hamilton2018}, to study compositionality in compounds from a diachronic viewpoint.

\section{Related Work}

From a synchronic perspective, \citet{reddy2011}, \citet{schulteimwalde2013} and \citet{schulteimwalde2016b} are closest to our approach, since they predict the compositionality of compounds using vector space representations. However, \citet{schulteimwalde2013} use German data and do not investigate diachronic changes. They report a Spearman's $\rho$ of 0.65 for predicting the compositionality of compounds based on the features of their semantic space and find that the modifiers mainly influence the compositionality of the whole compound, contrary to their expectation that the head should be the main source of influence. This is true for both the human annotation and their vector space model. \citet{schulteimwalde2016b} further investigate the role of heads and modifiers on the prediction of compositionality and report $\rho$ values between 0.35 and 0.61 for their models on German and English data. \citet{reddy2011} also report Spearman's $\rho$ between their surveyed compositionality values and word vectors. They achieve $\rho$ values of around 0.68, depending on the model.

From a diachronic perspective, we follow the general methodological approach of \citet{hamilton2018}, who use PPMI, SVD and word2vec based vector spaces to investigate a shift in meaning for chosen words with a known semantic change (\textit{gay}, \textit{broadcast}, etc.). They use time series to detect a significant change-point for two words, using cosine similarity and Spearman's $\rho$. They also compute the displacement for a single word embedding by calculating the cosine similarity between a point in time $t$ and a later point in time $t+\Delta$. We adapt this methodology and make use of the same corpus (Google Books Ngram).

\section{Methods and Data}
\label{sec:methods}

Several studies have been conducted in order to measure compositionality for compounds in different languages \cite{vonderheide2009,reddy2011, schulteimwalde2016a}. Some of these works have used large corpora to extract vector-based representations of compounds and their parts to automatically determine the compositionality of a given compound. The models were validated on the basis of their correlation with human compositionality ratings for a set of compounds. 

Because we are interested in the diachronic perspective on compounds, we use a time-stamped corpus: the Google Books Ngram corpus\footnote{The data is available from \url{https://commondatastorage.googleapis.com/books/syntactic-ngrams/index.html}} \citep{michel2010} It contains books from the 1500s to the 2000s, from which we retrieve the contextual information of compounds and their constituents per year. We operate on 5-grams, which is the largest unit provided by Google Ngrams and use the words appearing in the 5-grams as both target words and context. We use the Part-of-Speech information already included in the Google Ngram corpus to extract noun-noun patterns. We then regard all other tokens in the 5-gram as context words and from this build up a semantic space representation of noun compounds for each year. We use a sliding window approach, wherein we capture the context of a compound based on its position in the 5-gram. That means that a bigram (say the compound \textit{gold mine}) could occur in four different positions in the 5-grams (1-2, 2-3, 3-4 and finally 4-5). We then capture the contexts for each of these positions, in order to enrich the representation of a compound and its constituents (which similarly have five such positions, as they are unigrams).

Ideally, we would validate our diachronic model on diachronic test data. However, as it is not possible to survey compositionality rating for diachronic data, we instead use the synchronic data provided by \citet{reddy2011} (henceforth referred to as REDDY) for evaluating the quality of the Google Books Ngram data as a source for investigating the compositionality of compounds in general. \citet{reddy2011} compiled a list of 90 English compounds and asked annotators to rate the compositionality of these compounds on a scale from 0 to 5. They provide three mean values of their ratings for the compounds (\textit{compound-mean}), heads (\textit{head-mean}) and modifiers (\textit{modifier-mean}). We make use of REDDY in order to verify that our methods are capable of capturing compositionality (synchronically) and use the diachronic data of Google Books Ngram to investigate the temporal change of compositionality.

A common challenge in building semantic spaces on a diachronic scale is that when building the spaces for individual spans of time, the spaces need to be aligned later on in order to compare models \cite[see e.g.][Section 3.3]{kutuzov2018}. We circumvent this problem by jointly learning the spaces for the target words. To do this, we take the sparse representations of the compounds and their constituents and jointly learn their dense representations using SVD. Similar to \citet{hamilton2018} we also choose the dimensions of our embeddings to be 300. We carry out row normalization on the embeddings, in order to remove the bias of the frequency of the compounds and their constituents.

We make use of six different semantic features that have been proposed in the literature to capture compositionality \cite{schulteimwalde2016b} and  plausibility  of noun-noun compounds \cite{GuentherMarelli, dhar2019a}. Three features are based on the cosine similarity between the embeddings of different compound parts \citep[see][]{GuentherMarelli}:

\begin{enumerate}
    \item Similarity between compound constituents (\textit{sim-bw-constituents})
    \item Similarity of the compound with its head (\textit{sim-with-head})
    \item Similarity of the compound with its modifier (\textit{sim-with-mod})
\end{enumerate}

The three information theory based features given below were proposed by \citet{dhar2019a}: 

\begin{enumerate}
    \setcounter{enumi}{3}
    \item Log likelihood-ratio (\textit{LLR})
    \item Positive Pointwise Mutual Information (\textit{PPMI})
    \item Local Mutual Information (\textit{LMI})
\end{enumerate}

Such formulas have been used prior to calculate collocations and associations between words \citep[compare][]{manning}. Each feature will be tested individually for its ability to capture compositionality.

\section{Experiments}
\label{sec:exp}

We ran a total of two experiments\footnote{The code is available at \url{https://github.com/prajitdhar/Compounding}} (Section \ref{sec:exp1} and \ref{sec:exp2}) with different goals.

\subsection{Experimental Setup}
\label{sec:exp-setup}

\paragraph{Hyper-parameters} We experiment with certain hyper-parameters, in particular we varied the \textit{time span length}, e.g. single years, decades or a span of 20 years etc. and \textit{frequency cut-off} of compounds and their constituents in a specific time span, i.e. compounds and constituents have to occur above a certain frequency threshold. Choosing a greater time span will increase the observable data per compound and might improve the vector representations. We only consider compounds which retain representations in all time spans starting from the year 1800, which reduces the number of total compounds depending on the specific setup.

\paragraph{Compound-centric setting} \citet{dhar2019a} found the compound-centric set up, where the distributional representations of words are based on their usage as constituents in a compound to outperform compound-agnostic setups, for predicting novel compounds in English. They were inspired by research on N-N compounds in Dutch that suggests that constituents such as -molen `-mill' in pepermolen `peppermill' are separately stored as abstract combinatorial structures rather than understood on the basis of their independent constituents \cite{dejongetal2002}. We hence adopt the compound-centric setting. 

\subsection{Correlation}
\label{sec:exp1}

We first carry out a quantitative experiment, to see if our features bolster the prediction of compositionality in noun-noun compounds. To do so, we calculate correlation scores between our proposed metrics and the annotated compositionality ratings of REDDY. Like \citet{reddy2011} and \citet{schulteimwalde2013}, we opt for Spearman's $\rho$.

To find the best configuration of a time span and cut-off for the regression models, we use the $R^2$ metric. Table \ref{tab:temp_r2} presents our findings; we will discuss them in the following Section \ref{sec:results}.

\subsection{Progression of Compositionality over Time}
\label{sec:exp2}

Based on the results of our correlation experiment, we proceed to analyze the temporal progression of compositionality. Our goals are two-fold: First, investigate if temporal information helps in predicting the contemporary REDDY data and second, use the best feature and setup in order to model the progression of compositionality over time.

\section{Results}
\label{sec:results}

\begin{table}[htbp]
    \centering
    \begin{tabular}{@{}rrr@{}}
    \toprule
    Time span & Cut-off & R$^2$ $\pm$ sd \\
    \midrule
    \multirow{3}{*}{NA (Non-temporal)}
    & 20 & 0.343 $\pm$ 0.028 \\
    & 50 & 0.344 $\pm$ 0.026 \\
    & 100 & 0.337 $\pm$ 0.035 \\
    \midrule
    \multirow{3}{*}{1 (Year)}
    & 20 & 0.350 $\pm$ 0.029 \\
    & 50 & 0.171 $\pm$ 0.039 \\
    & 100 & 0.326 $\pm$ 0.030 \\
    \midrule
    \multirow{3}{*}{10 (Decade)}
    & 20 & 0.332 $\pm$ 0.024 \\
    & 50 & 0.328 $\pm$ 0.034 \\
    & 100 & 0.360 $\pm$ 0.062 \\
    \midrule
    \multirow{3}{*}{20 (Score)}
    & 20 & 0.341 $\pm$ 0.039 \\
    & 50 & 0.331 $\pm$ 0.031 \\
    & 100 & \textbf{0.370 $\pm$ 0.012} \\
    \midrule
    \multirow{3}{*}{50 (Half-century)}
    & 20 & 0.352 $\pm$ 0.038 \\
    & 50 & 0.360 $\pm$ 0.029 \\
    & 100 & 0.364 $\pm$ 0.034 \\
    \midrule
    \multirow{3}{*}{100 (Century)}
    & 20 & 0.351 $\pm$ 0.037 \\
    & 50 & 0.343 $\pm$ 0.033 \\
    & 100 & 0.344 $\pm$ 0.034 \\
    \bottomrule
    \end{tabular}
    \caption{R$^2$ values and standard deviation for the different configurations.}
    \label{tab:temp_r2}
\end{table}

We find the best predictors for the compositionality ratings of REDDY to be \textit{LMI} and \textit{LLR} (compare Table \ref{tab:reddycorr}). The overall highest correlation occurs between \textit{compound-mean} and \textit{LMI} with $\rho$ of 0.54. We also see that \textit{sim-bw-constituents} and \textit{sim-with-heads} are generally good predictors as well. Contrary to \citet{schulteimwalde2013} we do not find a strong correlation between modifiers and the REDDY ratings. Interestingly, \textit{PPMI} is always weakly negatively correlated with the ratings. This could be due to \textit{PPMI}'s property of inflating scores for rare events. As can also be seen from Table \ref{tab:reddycorr}, our correlation values are considerably lower than that of \citet{reddy2011}, but on par with a replication study by \citet{schulteimwalde2016b} for \textit{compound-mean}. We speculate that these differences are potentially due to the use of different data sets, the fact that we use a considerably smaller context window for constructing the word vectors (5 due to the restrictions of Google Ngram corpus vs. 100 in \citet{reddy2011} and 40 in  \citet{schulteimwalde2016a}) and the use of a compound-centric setting (as described in \ref{sec:exp-setup}).

\sisetup{detect-weight=true,detect-inline-weight=math}
\begin{table*}[htbp]
    \centering
    \begin{tabular}{@{}lSSS@{}}
    \toprule
    & \text{modifier-mean} & \text{head-mean} & \text{compound-mean} \\
    \midrule
    sim-bw-constituents & 0.35 & 0.41 & 0.48 \\
    sim-with-head & 0.26 & 0.43 & 0.43 \\
    sim-with-mod & 0.1 & 0.18 & 0.2 \\
    LLR & 0.36 & 0.44 & 0.52 \\
    PPMI & -0.12 & -0.1 & -0.14 \\
    LMI & \bfseries 0.38 & \bfseries 0.45 & \bfseries 0.54 \\
    \bottomrule
    \end{tabular}
    \caption{Spearman's $\rho$ of our measures and the compositionality ratings of REDDY.}
    \label{tab:reddycorr}
\end{table*}

From Table \ref{tab:temp_r2} we see that our best reported $R^2$ value occurs when observing time in stretches of 20 years (scores) and compounds having a frequency cut-off of at least 100. A few other observations could be made: In general the cut-off seems to improve the $R^2$ metric and the time spans of 10 and 20 years appear to be the most informative and stable for the cut-off values. Also, using temporal information almost always outperforms the setup that ignores all temporal information.

\begin{figure}[htbp]
    \centering
    \includegraphics[height=0.30\textheight]{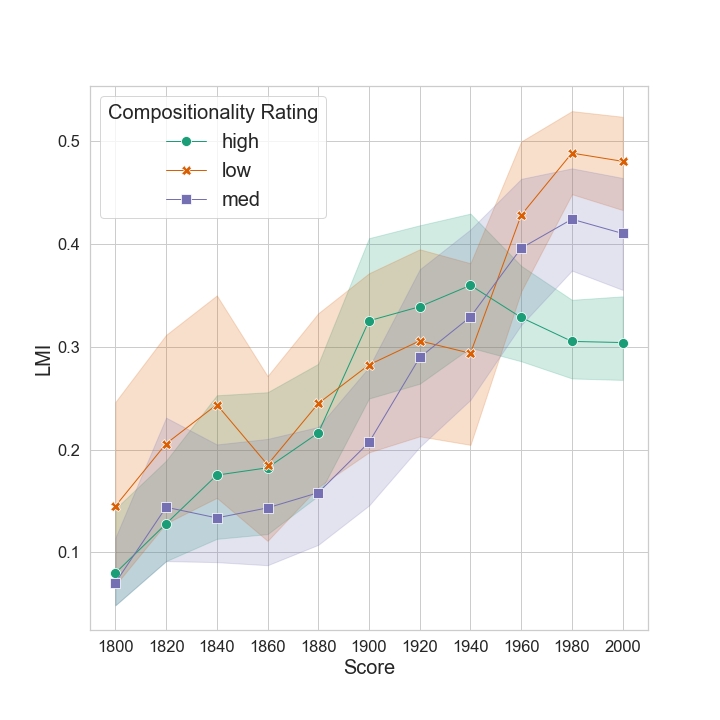}
    \caption{LMI of a compound in time point $t$ and $t+1$, with a time span of 20 years and a frequency cut-off of 100. Compounds are grouped according to their rating in REDDY.}
    \label{fig:lmi}
\end{figure}

For our following experiment, we choose to use the configuration with the highest $R^2$ value: a time span of 20 years and a cut-off of 100. Since LMI achieved the highest $\rho$ values, we also choose LMI over the other features. We group the compounds of REDDY into three groups based on the human ratings they obtained: \textit{low} (0-1), \textit{med} (2-3) and \textit{high} (4-5). Each group contains around 30 compounds. We then plot the LMI values of these three groups with their confidence interval across the time step of 20 years, shown in Figure~\ref{fig:lmi}. We can observe that there is a separation between the groups towards the later years, and that the period between 1940s and 1960s caused a noticeable change in the compositionality of the REDDY compounds. We find the same trends for all three information theory based features. Although care should be taken given the small data sets (especially for the earlier decades) on which the models were build and tested, the slope of the lines for the three groups of compounds seems to suggest that less compositional compounds go through a more pronounced change in compositionality than compositional compounds, as expected.

\begin{figure}[htbp]
    \centering
    \includegraphics[height=0.30\textheight]{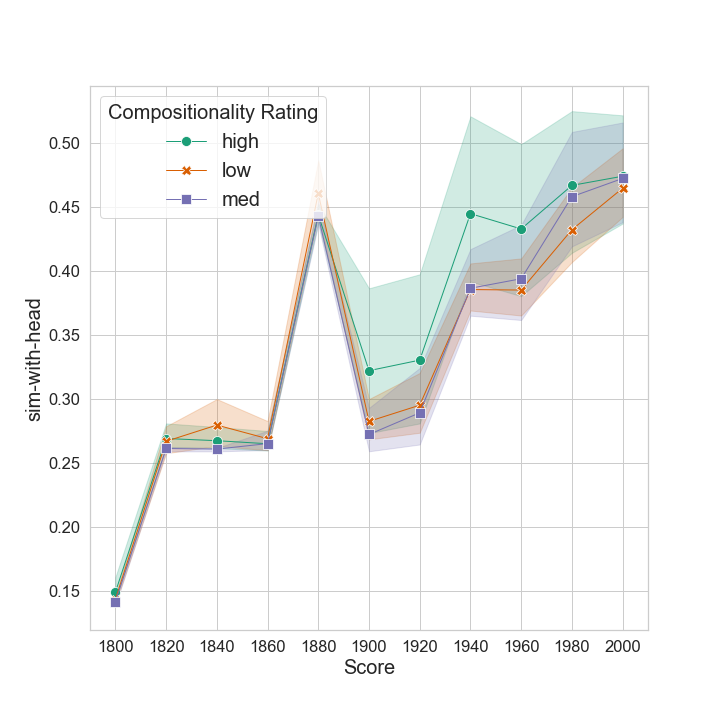}
    \caption{\textit{sim-with-head} of a compound in time point $t$ and $t+1$, with a time span of 20 years and a frequency cut-off of 100. Compounds are grouped according to their rating in REDDY.}
    \label{fig:sim-with-head}
\end{figure}

\begin{figure}[htbp]
    \centering
    \includegraphics[height=0.30\textheight]{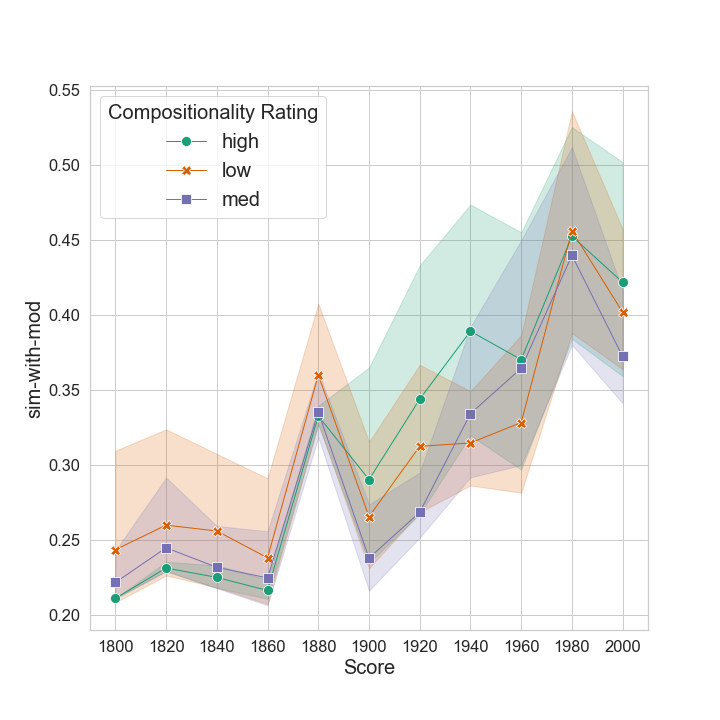}
    \caption{\textit{sim-with-mod} of a compound in time point $t$ and $t+1$, with a time span of 20 years and a frequency cut-off of 100. Compounds are grouped according to their rating in REDDY.}
    \label{fig:sim-with-mod}
\end{figure}

We also show the graphs for \textit{sim-with-head} and \textit{sim-with-mod} (Figures~\ref{fig:sim-with-head} and \ref{fig:sim-with-mod}) for the different groups of compounds across time, as these underperformed in our previous experiment. Both figures based on cosine based features largely confound the three groups of compounds across time, reinforcing our previous findings.

\section{Future Work}
Our current work was limited to English compounds from \citet{reddy2011}. We plan to expand our models to other languages for which compositionality ratings are available, such as German. 
We would also like to further investigate the fact that the information theory based measures outperform the ones based on cosine similarity. We intend to do so by incorporating more compounds and their compositionality ratings, as well as by using larger corpora.

Lastly, we will seek to find ways to harvest proxies for compositionality ratings of compounds over time. A possible avenue could be to use the information available in dictionaries.

\section{Conclusion}

We have shown work in progress on determining the compositionality of compounds over time. We showed that for our current setup, information theory based measures seem to capture compositionality better. Furthermore, we showed that adding temporal information increases the predictive power of these features to prognosticate synchronic compositionality. Finally, we showed how our best performing models trace the compositionality of compounds over time,  delineating the behavior of compounds of varying levels of compositionality.

\section*{Acknowledgements}

We would like to thank the anonymous reviewers for their valuable comments.
The second author has been funded by the Volks\-wagen~Foundation in the scope of the QuaDramA project.

\bibliography{acl2019}
\bibliographystyle{acl_natbib}


\end{document}